\documentclass[11pt,a4paper]{article}

\usepackage{indentfirst}
\usepackage{geometry} 
\usepackage{graphicx}
\usepackage{subfigure}

\title{A Deep Reinforcement Learning Strategy for UAV Autonomous Landing  on a  Platform}
\author{Z. Jiang$^{1}$, G. Song$^{1}$\\
$^{1}$School of Aerospace and Astronautics, Zhejiang University}
\date{}

\begin{document}
\maketitle

\noindent
\textbf{Abstract    }

With the development of industry, drones are appearing in various field. In recent years, deep reinforcement learning has made impressive gains in games, and we are committed to applying deep reinforcement learning algorithms to the field of robotics, moving reinforcement learning algorithms from game scenarios to real-world application scenarios. We are inspired by the LunarLander of OpenAI Gym, we decided to make a bold attempt in the field of reinforcement learning to control drones. At present, there is still a lack of work applying reinforcement learning algorithms to robot control, the physical simulation platform related to robot control is only suitable for the verification of classical algorithms, and is not suitable for accessing reinforcement learning algorithms for the training. In this paper, we will face this problem, bridging the gap between physical simulation platforms and intelligent agent, connecting intelligent agents to a physical simulation platform, allowing agents to learn and complete drone flight tasks in a simulator that approximates the real world. We proposed a reinforcement learning framework based on Gazebo that is a kind of physical simulation platform (ROS-RL), and used three continuous action space reinforcement learning algorithms in the framework to dealing with the problem of autonomous landing of drones. Experiments show the effectiveness of the algorithm, the task of autonomous landing of drones based on reinforcement learning achieved full success.


\section{Introduction}
In the recent years, the use of drones has become increasingly popular. multi-rotor drones demonstrates its great potential in many scenarios, from drone disaster, relief logistics and warehousing to a wide range of automated industries. However, these tasks are achieved by a relatively independent group of key components, each component is independent and complex. Researchers usually decompose these components to study. Nearly every task has the component of autonomous landing of drones [10][11]. Classic technology landing techniques have their many limitations, such as  design of the model, nonlinear approximation, as well as anti-interference and effective computational aspects. And these classical technologies of autonomous landing have their great limitations, one algorithm can be only applied in one drone model, such as PID, each model depends on a special tuple of PID parameters that relies on extensive manual experience[10].

In this area, machine learning technique proved to be an effective means of overcoming difficulties, machine learning has evolved into many complex application areas[9]. While traditional reinforcement learning is mainly popular in the field of games, and has less application in our real life.We want to apply reinforcement learning methods in machine learning to the field of drones, using reinforcement learning to train drones to achieve autonomous landing tasks. There are many problems with training intelligent agents of drones in the real world, on the one hand training agents in the real world is unsafe, on the other hand, we need a lots of drones, the cost is huge during the training process. But the physical simulation platform can help us to overcome these difficulties. In order to make sure the authenticity of simulation, we choose Gazebo as our simulation platform, Gazebo is a very powerful simulator, it often be used to simulate robotic arm and other physical robotics, with the help of the platform ,we can realization of motion simulation of drone. At the same time, PX4 is prefer to recommend Gazebo as simulation simulator. It follows that the performance of Gazebo is reliable. We choose Gazebo as the simulation platform here and perform reinforcement learning training of agents in Gazebo environment.Reinforcement learning algorithms can be divided into continuous action space algorithms and discrete action space algorithms according to the action space. Because the control of drones is continuous control, we choose to use the continuous action space algorithm for training. In this paper, continuous action space reinforcement learning algorithm, DDPG TD3 SAC, are applied. 

In our work, we used reinforcement learning (DDPG,TD3,SAC) to train drones to achieve autonomous landing and this task is deployed in our reinforcement learning framework based on Gazebo. Using Gazebo Simulator and PX4 Emulator during the training process. Rodriguez-Ramos used RotorS[13]as simulation framework to deal with autonomous landing tasks[2],RotorS is only a simulator, PX4 can not only support simulation but also live deployment, so PX4 considers more flight details.

The remainder of the paper is organized as follows:Section 2 presents a brief introduction on the reinforcement
learning theory and brief explains the basics of TD3 algorithm and SAC algorithm. Section 3 details the presentation and
description of our Gazebo-based reinforcement learning
framework(ROS-RL) and how to design scenarios and tasks for drones autonomous landings in Gazebo, then details the description of the reinforcement learning algorithms combining with the tasks. In section 4, some experiment results show the availability of our Gazebo-based reinforcement learning
framework(ROS-RL) and comparative analysis of the performance of several algorithms used in the experiments based on the experimental results. 

\section{Background}
In reinforcement learning, agents are used to interact with the environment, an agent gets a state in the environment, an then uses this state to output an action, after that this action will be applied in the environment, environment will output the next state and the reward according to the current action. The goal of the agent is try to collect reward from the environment as much as possible. 

During the reinforcement learning process, agents constant interact with the environment in states and actions, adjust neural network parameters based on reward feedback from the environment, finally, the agents learn a behavioral strategy that can result a high reward from the environment. This is a process of adjusting the neural network parameters, and in this process of optimizing the parameters and finding the optimal solution, agents need to balance exploration and exploitation. Exploration means there is a certain probability of not following the decision made by the agent in the beginning of training, exploring the environment randomly or in other ways to find a better solution. The increase in exploration probability is more conducive to finding a better solution, but against the sensitivity of decision making, and affects training efficiency. agents only balance the problem of exploration-exploitation to find the optimal strategy and the maximum cumulative reward in the process of interacting with the environment.

Reinforcement learning consists of three parts--policy, value function, model. Policies can map the state of each environment to an action in the action space, they can be divided into two groups, stochastic policy and deterministic policy. Stochastic policy is $ \pi $, $\pi (a|s)=p(a_{t}=a|s_{t}=s )$, inputs a status $s$, outputs a series of probabilities, these probabilities represent probability of each action of the agent. Deterministic policy, $a^{*}= arg_a max \pi (a|s)$, outputs the action has the highest probability. There are two kinds of value functions, one kind of value function is $V_\pi(s)$, the other kind of value function is $Q_\pi(s,a)$, The first kind of value function $V_\pi(s)=E_\pi[G_t|s_t=s]=E_\pi[\sum_{k=0 }^{\infty}\Upsilon ^{k}r_{t+k+1}|s_t=s ]$means when we use policy $\pi$, how much value the agent can accumulate from status $s$ to the end status; the second value function can be called $Q$ function,  $\mathrm{Q}_{\pi}(\mathrm{s}, \mathrm{a}) = E_{\pi}\left[\mathrm{G}_{\mathrm{t}} \mid \mathrm{S}_{\mathrm{t}}=\mathrm{s}, \mathrm{A}_{\mathrm{t}}=\mathrm{a}\right]=E_{\pi}\left[\sum_{\mathrm{k}=0}^{\infty} \gamma^{\mathrm{k}} \mathrm{R}_{\mathrm{t}+\mathrm{k}+1} \mid \mathrm{S}_{\mathrm{t}}=\mathrm{s}, \mathrm{A}_{\mathrm{t}}=\mathrm{a}\right]$, this value function use the current state and current action to estimate expectations for future rewards. This way of calculating the value function is used in many network structures such as TD3. The third part is the model, the model determines the state of the next step. The state of the next step depends on the current state and the current action chosen. The state transfer probabilities and reward functions of the model are generally defined by the following form, transfer probabilities $\mathcal{P}_{\mathrm{ss}^{\prime}}^{\mathrm{a}}=P\left[\mathrm{S}_{\mathrm{t}+1}=\mathrm{s}^{\prime} \mid \mathrm{S}_{\mathrm{t}}=\mathrm{s}, \mathrm{A}_{\mathrm{t}}=\mathrm{a}\right]$, reward function $\mathcal{R}_{\mathrm{s}}^{\mathrm{a}}=E\left[\mathrm{R}_{\mathrm{t}+1} \mid \mathrm{S}_{\mathrm{t}}=\mathrm{s}, \mathrm{A}_{\mathrm{t}}=\mathrm{a}\right]$.

Now, we have the three elements of strategy, value function and model, we can implement the interaction with the environment and the collection of experience to train our neural network model.

Generally speaking, our drone problem is a continuous state space problem. Because continuous states and actions generate a large action space and state space,which makes neural network optimization more difficult. Traditional reinforcement learning has the limitation of small state space and small action space, and is only applicable to problems in discrete state space. The addition of deep neural networks to reinforcement learning allows these problems to be solved effectively,TD3 SAC are two successful deep reinforcement learning algorithms. 

The earliest continuous action space reinforcement learning algorithm would be DDPG, TD3 is further improved and upgraded on the basis of DDPG, it is an Actor-Critic algorithm, it can effectively solve the problem of overestimation of DDPG and make the training process more stable, new noise exploration mechanisms are combined in TD3. TD3 designs two sets of Critic networks to solve the problem of over-evaluation of the value function $Q$, two sets of networks are evaluated for behaviors separately, and finally a lower Q value is taken as the final Q value, and then an Actor network is used to make behavioral decisions. At the same time, there are also two Gaussian noises in the network, a Gaussian noise is added to the action vector of Q(s,a) when the Critic network evaluates the Q value, and a Gaussian noise is added to the original action after the action generated by Actor. Gaussian noises help to balance the problem of exploration-exploitation.Adding gradient intercept to avoid the gradient of parameter update out of a certain range. Finally, Critic and Actor use different update frequencies to improve the stability of training. Updating network parameters uses soft update, $\theta=\tau \theta^{\prime}+(1-\tau) \theta$, instead of copy directly. 

SAC is also an Actor-Critic algorithm, it uses both kinds of value function models $V_\pi(s)$ and $Q_\pi(s,a)$ mentioned above to evaluate the status value more accurate. On the other hand, action entropy is added in the value function $V_\pi(s)$, action entropy encourages  Actor to explore toward more random directions. In this way, convergence time of the algorithm has been significantly reduced. The ideas of two sets of Q-Critic network and soft update in TD3 algorithm are still applied in SAC. We share more details in Section 3.4.

\section{Proposed Method}
In this section, we present a complete explanation of our Gazebo-based
reinforcement learning framework(ROS-RL). And using a case study to demonstrate the effectiveness of our framework, an agent can be able to learn and control a drone for autonomous landing tasks using this framework. In this framework, agent finds out the effective strategy to manipulate the drone to land on the parking platform through many failures. In this section we describe more details about these.
\subsection{Reinforcement Learning Framework}
Reinforcement learning algorithms have been well tested and validated in atari and other games. But in some complex continuous state space and action space, there are still many question about how it will perform in the physical continuous scenarios. However, the need to apply reinforcement learning to real physical spaces is becoming increasingly urgent. Gazebo simulator, as the first choice simulator for simulating real robots, is widely used in the field of robotics[3][4][5]and it is supported by the famous plug--Robot Operating System(ROS). The algorithm validated by ROS-Gazebo can be easily transplant to realistic robot system controlled by ROS. So we designed a Gazebo-based reinforcement learning
framework, and it can build up the bridge between the reinforcement learning algorithm and  physical simulation scenarios in Gazebo. It helps us to train our reinforcement learning algorithms cost-effectively and safely, and to observe how well the algorithms perform roughly when deployed them into the physical real world. 

Reinforcement learning framework mainly composed of Gazebo simulator, ROS plugs, environment interface and agents. We can arrange robots in Gazebo to simulate our physical real world. We can arrange drones to Gazebo, add landmarks, obstacles and other environmental elements and configure their materials, density, surface friction, collision elasticity and other kinds of detailed content. After that is ROS plugs, ROS framework is a popular robot control system in the field of robotics, it works as a node, ROS makes it easy to interact with the elements in the Gazebo world, include drones' status and modify the parameters about the drones. Environment interface is used to encapsulate data coming from ROS and extract the valuable data for the agent, in the same way, we only need to submit the fundamental parameters when we want to control the drones in the gazebo, other parameters will be filled automatically by the environment interface. The last part is agents, various kind of algorithms belong to this part. Agents collect the environment status and rewards to adjust the policy network, and output actions to collect more rewards. The basic structure of framework has been show as Fig.1.
\begin{figure}[htbp]
    \centering
    \includegraphics[scale=0.6]{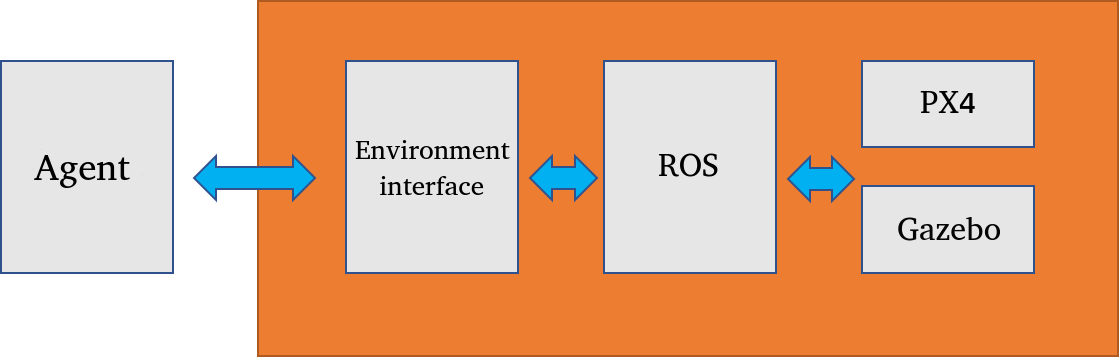}
    \caption{The reinforcement learning framework based on Gazebo(ROS-RL), Gazebo simulator, ROS plug, environment interface, and agents. Reinforcement learning environment highlighting with red background}
    \label{fig:图片标签1}
\end{figure}

\subsection{Tasks And Scenarios Design}
After introducing the framework, we inject the reinforcement learning algorithm into the framework and arrange a simple environment(which includes a drone[Fig.2] from PX4 model library and a landmark for landing[Fig.3]), then use the framework to training the agent and make it learn to control drone to complete the landing task. In this task, the drone needs to reasonably adjust its flight speed and flight attitude according to the self-information in the environment. Landing on the 
landmark in a limited time step. 
\begin{figure}[htbp]
    \centering
    \begin{minipage}[t]{0.48\textwidth}
        \centering
        \includegraphics[width=6cm]{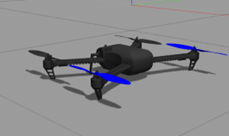}
        \caption{the drone models}
    \end{minipage}
    \begin{minipage}[t]{0.48\textwidth}
        \centering
        \includegraphics[width=6cm]{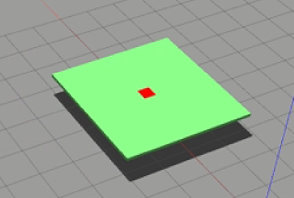}
        \caption{landing landmarks}
    \end{minipage}
\end{figure}

In this task, the drone is using classic model iris in PX4 models library. Landmark is a  4m\*4m\*0.1m box, Landmarks consist of two types of areas, external area is a 4m\*4m green floor and middle area is a 0.5m\*0.5m red floor, landing on the red area gets a higher reward than the green area. Because the PX4 drone uses closed-loop simulation, flight accidents from collisions and other causes during flight can lead to irreversible failure problems. To reduce the trouble of resetting the model, we removed the collision property of the landing landmark.  The height of the upper surface of the landmark is 1m, the drone is considered to have landed on the landmark when the drone lands at a height of 1m. Each episode the drone starts landing at a height of 3m near the landmark, The mission ends when the drone is below an altitude of 1m or after finishing 40 time steps. At the end of the mission, the UAV automatically returns to an area at an altitude of 3m near the landmark and prepares for the next landing mission. 

\subsection{Reinforcement Learning Problem Definition And Function Design}
In the reinforcement learning, The tuples of experience plays a decisive role in convergence of algorithm. The design of status space, action space and reward value function affect the convergence of the algorithm and the speed of convergence directly. 

The tuples of experience in reinforcement learning are consisted by status space $ s \in S $, action space $a \in A$ and reward function $ r $. In our experiment, we defined the status space$S$ like as Eqs.1.

$S=\left\{p_{x}, p_{y}, p_{z}, v_{x}, v_{y}, v_{z}\right\}$				(1)

In this equation, $p_x,p_y,p_z$ represent the offset of the drone relative to the landing landmark on the x-axis, y-axis and z-axis at the current moment. $v_x,v_y,v_z$ represent the velocity of the drone relative to the landing landmark. These row data reads from the Gazebo simulation via the ROS interface and then encapsulates by environment interface for the agent. The action space $A$ is defined by Eqs.2. 

$A=\left\{a_{x}, a_{y}\right\}$				(2)

In this equation, $a_{x}, a_{y}$ represent the velocity of the drone relative to the ground on the x-axis, y-axis.Agent generates $a_{x}, a_{y}$, and submit them to the FCU of drone in the Gazebo through environment interface and ROS plug.The drone generates a certain pitch and roll angle under the control of FCU. In this way, the drone generates a certain horizontal velocity on the x-axis and y-axes. Although the agent is designed to solve the problem of three-dimensional continuum of action space. The appearance of $a_z$ will make the solution space of the agent very large, it is not friendly to the convergence of the algorithm. Therefore the velocity of the drone in the z-axis direction is not directly generated by the agent. It is generated by the linear equation  $a_z=\alpha*p_z$, $\alpha$ is a hyper-parameter, the lower the altitude of the drone the slower the landing speed. At the same time, the landing speed of the drone in the z-axis is also related to the flight attitude of the drone, affected by $a_{x}, a_{y}$ indirectly. Generally speaking, this landing task consists of a six-dimensional continuous state space and a two-dimensional continuous action space. 

After that, we introduce the design of the reward function. Reward function is the most important function in reinforcement learning, it affects the convergence of the algorithm and the speed of convergence directly. When agent receives a $s$, and outputs an $a$ for the environment, environment feedback a $s$ and a reward $r$. Agent can get a positive reward in each step if agent  gets closer to the target. Reward function is defined by Eqs.3. and Eqs.4. 

$ shaping_{t}=-100 \sqrt{p_{x}^{2}+p_{y}^{2}+p_{z}^{2}}-10 \sqrt{v_{x}^{2}+v_{y}^{2}+v_{z}^{2}}-\sqrt{a_{x}^{2}+a_{y}^{2}} +10 C\left(1-\left|a_{x}\right|\right)+10 C\left(1-\left|a_{y}\right|\right)$       (3)

$r=shaping_{t}-shaping_{t-1}$				(4)

Eqs.3 reflects the importance weighting of each evaluation indicator to task completion, $p_{x}, p_{y}, p_{z}$ is the relative position of the drone with respect to the landing landmark, it is also one of the most important evaluation indicators during the landing of drones, they are given the highest weight, $v_{x},v_{y},v_{z}$ are used to evaluate the speed of the drone relative to the landing landmark during the landing process, these speeds determine whether the drone can land smoothly on top of the landmark, the design of $\sqrt{v_{x}^{2}+v_{y}^{2}+v_{z}^{2}}$ promotes a smooth landing process for the drone. $a_{x}, a_{y}$ are the amount of control over the movements of the drone, $\sqrt{a_{x}^{2}+a_{y}^{2}}$ promotes that the agent can output the control speed of each time step more smoothly. $C$ is a hyper-parameter used to indicate whether the drone landed on the ground target. $C$ keeps a 0 value if the drone did not touch the ground target, when the drone lands on the ground target, agent will get a bonus based on the amount of throttle on the x-axis y-axis. We expect the amount of throttle on the x-axis y-axis tend to zero when the drone landing on the ground target. For this reason, agent can get a biggest reward if $a_x,a_y$ tends to zero at the moment the drone hits the ground. Finally, discussing the design of shaping in Eqs.4, shaping is a popular design approach to accelerate the convergence of reinforcement learning. It improves the convergence speed of reinforcement learning algorithms by passing information about the current progress of the task to be solved shaping.[1]
\subsection{Reinforcement Learning Algorithm And Network Structure Design}
In order to solve the problem of controlling the drone to land autonomously by reinforcement learning in the continuous state space and continuous action space. We conducted comparative experiments using various reinforcement learning algorithms, it includes DDPG[6], TD3[7]and SAC[8].
\subsubsection{DDPG}
DDPG is an improved algorithm based on the DQN algorithm. DQN gives up the Q-Table in the QLearning, and uses the neural network instead of the Q-table to deal with continuous states, but the action is still discrete. DDPG adds an actor to the DQN, which makes DDPG can output continuous action. And DDPG can deal with some problems with continuous action and continuous state. The purpose of the Actor is to output an appropriate action according to the state and the algorithm hopes agent can get bigger Q-value after putting the action into Critic network. So DDPG is an Actor-Critic framework, Actor learns policy function and improves the possibilities of action that can gets highest reward, Critic network learns action value function and try to evaluate the real reward for the action. DDPG inherits the idea of DQN and uses a two-group network structure, one group plays the learning network(parameters $Q,u$) and it needs to update the parameters frequently, another group network plays as target network(parameters $Q',u'$) and updates the parameters of the network with  a lower frequency. The loss of the Q-network is calculated as Eqs.5.

$ Loss =\left(Q\left(s_{t}, a_{t}\right)-\left(r_{t}+Q\left(s_{t+1}, \mu\left(s_{t+1}\right)\right)\right)\right)^{2}$				(5)

Each update changes the Q-network, which makes the network to be difficult to converge, target network(parameters $Q',u'$) comes up to deal with this problem. The loss of the Q-network is calculated as Eqs.6.

$Loss =\left(Q\left(s_{t}, a_{t}\right)-\left(r_{t}+Q'\left(s_{t+1}, \mu'\left(s_{t+1}\right)\right)\right)\right)^{2}$				(6)

In this equation, only the parameters of the Q network are changing, which can improve the convergence speed of the network. Actor updates parameters with gradient ascent method(Eqs.7.).

$\Delta \theta_{\mu}=\nabla_{\theta_{\mu}} Q(s, a)=\nabla_{a} Q(s, a) \nabla_{\theta_{\mu}} a=\left.\nabla_{a} Q(s, a)\right|_{a=\mu(s)} \nabla_{\theta_{\mu}} \mu(s)$				(7)

DDPG uses the soft-update method to update the parameters in the network, not copies the parameters of learning network to the target network but do a few adjustment on the original network(Eqs.8 and Eqs.9). 

$\theta_{Q^{\prime}}=\tau \theta_{Q}+(1-\tau) \theta_{Q^{\prime}}$				(8)          
$\theta_{\mu^{\prime}}=\tau \theta_{\mu}+(1-\tau) \theta_{\mu^{\prime}}$				(9)
\subsubsection{TD3}
 TD3 is an improved algorithm based on the DDPG. TD3 still uses the Actor-Critic structure in DDPG. TD3 designs two sets of Critic networks to relieve the Q-value overestimate problem in DDPG. Two sets of Critic network evaluate the Q-values at the same time and choose the smaller Q-value as valid value to ease overestimate. Besides, the action outputed by the Actor is added some noise before evaluating the Q-value. After several updates, Critic network can evaluate the Q-value more accurate with the help of noise. 

$ Loss _1=\left(Q_1\left(s_{t}, a_{t}\right)-\left(r_{t}+min_{i=1,2}Q_i'\left(s_{t+1}, \mu'\left(s_{t+1}\right)+noise\right)\right)\right)^{2}$				(10)

$ Loss _2=\left(Q_2\left(s_{t}, a_{t}\right)-\left(r_{t}+min_{i=1,2}Q_i'\left(s_{t+1}, \mu'\left(s_{t+1}\right)+noise\right)\right)\right)^{2}$				(11)

Updating the parameters of Actor network depending on the Critic network, because keeping the same frequency of updates is not good for Actor network to find out the optimal strategy. So TD3 uses delayed update strategy, only when Critic network is more stabilized, Actor will update its network. The gradient of the Actor update is generated in the same way as the DDPG algorithm. TD3 has the same soft-update method as DDPG. 
\subsubsection{SAC}
SAC is also a reinforcement learning algorithm based on the Actor-Critic structural, the difference from other algorithms is that SAC tries to maximize the entropy of policy, which makes the strategy as random as possible and makes the agent to fully explore the state space. SAC performs well in a real robot control task.

The SAC algorithm is more complex than the first two algorithms. It is consisted by an Actor network and two V-Critic(a learning network and a target network) networks and two Q-networks(two learning networks). The actor network inputs the environment states and outputs the probability distribution of the action. V-Critic inputs the status of environment and Q-Critic inputs the status of environment and actions. Three kinds of networks use different parameter update methods. For Q-network updates, calculate the value of $t+1$ step $V(s_{t+1})$ through V-Critic network firstly, and then sample key-Value pairs in which the action $a_t$ can transfer the status $s_t$ to status $s_{t+1}$ from the experience pool. Calculate the Q-value $Q(s_t,a_t)$ at $t$ step, ideally it would satisfy the Eqs.12. 

$Q(s_t,a_t)=r_{t+1}+\gamma V(s_{t+1})$				(12)

In reality, there are some deviations in network prediction. The left and right sides of the equation cannot be equal, the left side of the equation is closer to the real value, the value gap between the left and right sides of the equation is the Loss of Critic network. Entropy is introduced in V-Critic. Entropy is used to describe the degree of chaos, and it is defined by the Eqs.13. 

$ H=-\sum_{x} p(x) \log p(x)=\sum_{x} p(x) \log \frac{1}{p(x)} $				(13)

The updates of V-Critic depends on the Q-Critic network, Q- Critic evaluates the value of status $s_t$ with the equation $E_{a_{t}^{\prime} \sim \pi\left(\cdot \mid s_{t} ; \theta\right)}\left[\min _{i=0,1} q_{i}\left(s_{t}, a_{t}^{\prime} \right)\right] \approx  V(S_t)$, $V(S_t)$ adds an item of entropy ($ H=-\sum_{x} p(x) \log p(x)$) to help the V-Critic explores the space more fully. The expectation of $V(S_t)$ is Eqs.14. $\alpha$is a coefficient indicates the importance.

$V(S_t) =  E_{a_{t}^{\prime} \sim \pi\left(\cdot \mid s_{t} ; \theta\right)}\left[\min _{i=0,1} q_{i}\left(s_{t}, a_{t}^{\prime} \right)-\alpha \ln \pi\left(a_{t}^{\prime} \mid s_{t} ; \theta\right)\right]$				(14)

The higher the entropy of the action generated by the policy network n Eqs.14, the value $V(S_t)$ higher. Therefore, the entropy in V-Critic can encourage the agent to explore the space. The real value can be calculated by the right side of Eqs.14. Left side of Eqs.14 is the predicted value of V-Critic network, the difference between left and right is the loss of V- Critic. Actor updates the parameters using the gradient ascent method, $ \Delta \theta =\nabla_{\theta}E_{a_{t}^{\prime} \sim \pi\left(\cdot \mid s_{t} ; \theta\right)}\left[q_{0}\left(s_{t}, a_{t}^{\prime} \right)-\alpha \ln \pi\left(a_{t}^{\prime} \mid s_{t} ; \theta\right)\right] $. SAC uses soft-update method to refresh the parameters from V-Critic learning network to V-Critic target network  , $\theta_{V^{\prime}}=\tau \theta_{V}+(1-\tau) \theta_{V^{\prime}} \quad $.
\section{Experiment}
In this section, we will present the design details of the experiment include the environmental platform for experiments, construction and design of simulation scenarios and training process of drone. Finally, the results of the experiment are summarized and discussed. 
\subsection{Experimental configuration introduction}
The simulation platform builds up on Ubuntu 20.04 and using ROS neotic. Simulation environment using Gazebo11, PX4 as flight control software. Reinforcement learning neural network constructed and trained using torch 1.11.0 and accelerated training process with Nvidia GTX 1050 Ti
\subsection{Simulation scenarios introduction}
The simulation scenario was created in Gazebo 11 and the scenario contains a UAV model and a landing platform (see Fig.4).
\begin{figure}[htbp]
    \centering
    \includegraphics[scale=0.6]{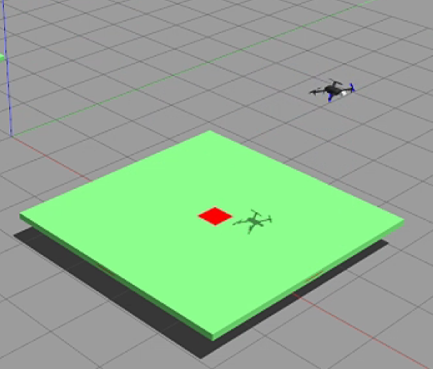}
    \caption{Simulation scenarios, drone and landing platform in the training process}
    \label{fig:图片标签4}
\end{figure}
The drone comes from PX4 open-source model library, PX4 prepared various kinds of drone models and sensors include IMU, depth cameras, RGB cameras and so on. We chose the general model iris to accomplish the autonomous landing mission. In order to realise the interaction between the agent and simulation scenarios, we read and launch relevant data(the flight status of the drone and the control commands of the drone) through the environment interface, environment interface to the virtual world through the ROS plug, the raw data of the virtual world is obtained through ROS topics, and then the valid data is extracted and encapsulated for submission to the experience pool of the agent. In the same way, control commanders are un-encapsulated and submit to corresponding topics about drones controlling through ROS plug. The entire interactive training process takes place in Gazebo, interaction is performed at a frequency of 10Hz.
\subsection{Experience Results And Discussion}
In this section, the above three algorithms are validated to check their effectiveness in controlling the landing of drones. Then the reinforcement learning algorithm is evaluated in terms of training time and control effectiveness of the drones. 
In experiments, we trained the three reinforcement learning algorithms for more than 1000 episodes under the same conditions. The reward value curves of three algorithms are shown in the Fig.5, Fig.6, Fig.7. Experimental results show that DDPG unable to converge, can not solve the problem of controlling the autonomous landing of drones in a 3D environment. TD3 can complete the task of controlling the autonomous landing of the drone through training, TD3 has good convergence, but needs long training time and the exploration of space is insufficient. We can find out in Fig.6 that TD3 can complete autonomous landing of the drone after 1200 episode of training and the reward value reaches its highest point, because of the noise in TD3 algorithm, the drone landing is not smooth, the smoothness evaluation term in the reward function is affected, the convergence of the reward value is lower than the SAC algorithm. In the SAC algorithm, entropy is considered to encourage the agent to explore fully in the space. In this way, SAC reaches convergence in a shorter time, TD3 algorithm needs 1200 episodes to learn the strategy of landing, but SAC algorithm only need 300 episodes and its reward can reach a higher point. SAC can learn the landing strategy of the drone in a shorter time. But SAC uses the idea of maximum entropy to encourage exploration, and the algorithm is still in high intensity exploration after the convergence. Which results in a convergence curve that is not as stable as TD3 after 300 episodes. Optimizing the annealing method of the weight $\alpha$ before the entropy term can improve the problem of unstable convergence.  
\begin{figure}[htbp]
    \centering
    \begin{minipage}[t]{0.3\textwidth}
        \centering
        \includegraphics[width=5cm]{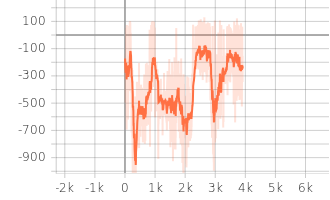}
        \caption{Reward per episode of DDPG}
    \end{minipage}
    \begin{minipage}[t]{0.3\textwidth}
        \centering
        \includegraphics[width=5cm]{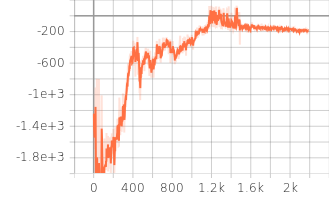}
        \caption{Reward per episode of TD3}
    \end{minipage}
        \begin{minipage}[t]{0.3\textwidth}
        \centering
        \includegraphics[width=5cm]{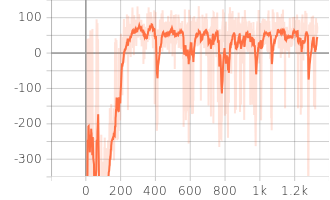}
        \caption{Reward per episode of SAC}
    \end{minipage}
\end{figure}
\section{Conclusions And Future Work}
In this paper, we created a reinforcement learning
framework based on Gazebo simulator and used the ROS as data interaction interfaces, follow by that, we use reinforcement learning to solve the problem of autonomous drone landing. We accessed three continuous action space reinforcement learning algorithms to the ROS-RL framework. Training and comparing the effectiveness of three algorithms to control the autonomous landing of drones. The results show that two of the three algorithms successfully learned the strategy to control the autonomous landing of the drones after training and the two algorithms have advantages and disadvantages in terms of convergence stability and convergence time of the training process. In the nearly future, we can apply this framework to more complex scenarios and use reinforcement learning to allow agents to learn solution strategies of other complex tasks in a simulation environment. At the same time, ROS as a popular interface for physical robot control, our simulation framework uses the ROS interface for training. It is relatively easy to transfer the algorithms trained in the virtual simulation scenario to the real scenario. We will explore the "Simulation training, Live deployment 
" solutions[12], make the maximize benefits of ROS-RL.

\newpage
\section{Reference}

[1]Dorigo M, Colombetti M. Robot shaping: an experiment in behavior engineering[M]. MIT press, 1998.

[2]Rodriguez-Ramos A, Sampedro C, Bavle H, et al. A deep reinforcement learning strategy for UAV autonomous landing on a moving platform[J]. Journal of Intelligent \& Robotic Systems, 2019, 93(1): 351-366.

[3]Polvara R, Patacchiola M, Sharma S, et al. Toward end-to-end control for UAV autonomous landing via deep reinforcement learning[C]//2018 International conference on unmanned aircraft systems (ICUAS). IEEE, 2018: 115-123.

[4]Imanberdiyev N, Fu C, Kayacan E, et al. Autonomous navigation of UAV by using real-time model-based reinforcement learning[C]//2016 14th international conference on control, automation, robotics and vision (ICARCV). IEEE, 2016: 1-6.

[5]Yu X, Fan Y, Xu S, et al. A self‐adaptive SAC‐PID control approach based on reinforcement learning for mobile robots[J]. International Journal of Robust and Nonlinear Control, 2021.

[6]Lillicrap T P, Hunt J J, Pritzel A, et al. Continuous control with deep reinforcement learning[J]. arXiv preprint arXiv:1509.02971, 2015.

[7]Fujimoto S, Hoof H, Meger D. Addressing function approximation error in actor-critic methods[C]//International conference on machine learning. PMLR, 2018: 1587-1596.

[8]Haarnoja T, Zhou A, Abbeel P, et al. Soft actor-critic: Off-policy maximum entropy deep reinforcement learning with a stochastic actor[C]//International conference on machine learning. PMLR, 2018: 1861-1870.

[9]Janousek J, Marcon P, Klouda J, et al. Deep Neural Network for Precision Landing and Variable Flight Planning of Autonomous UAV[C]//2021 Photonics \& Electromagnetics Research Symposium (PIERS). IEEE, 2021: 2243-2247.

[10]Moriarty P, Sheehy R, Doody P. Neural networks to aid the autonomous landing of a UAV on a ship[C]//2017 28th Irish Signals and Systems Conference (ISSC). IEEE, 2017: 1-4.

[11]Ciabatti G, Daftry S, Capobianco R. Autonomous Planetary Landing via Deep Reinforcement Learning and Transfer Learning[C]//Proceedings of the IEEE/CVF Conference on Computer Vision and Pattern Recognition. 2021: 2031-2038.

[12]Zhao T, Jiang H. Landing system for AR. Drone 2.0 using onboard camera and ROS[C]//2016 IEEE Chinese Guidance, Navigation and Control Conference (CGNCC). IEEE, 2016: 1098-1102.

[13]Furrer F, Burri M, Achtelik M, et al. RotorS—A modular gazebo MAV simulator framework[M]//Robot operating system (ROS). Springer, Cham, 2016: 595-625.

\end{document}